\pdfoutput=1

\documentclass[11pt]{article}

\usepackage{emnlp2023}

\usepackage{times}
\usepackage{latexsym}
\usepackage{graphicx}
\usepackage{booktabs}
\usepackage{fancyvrb}
\usepackage{enumerate}
\usepackage[T1]{fontenc}
\usepackage{colortbl}
\usepackage{color}
\usepackage{soul}

\usepackage[utf8]{inputenc}
\usepackage{amsmath}
\usepackage{multirow}
\usepackage{array}
\usepackage{siunitx}
\usepackage{booktabs}
\usepackage{comment}

\usepackage{microtype}

\newcommand{\overlap}{QUIP-Score}
\newcommand{\overlaplong}{QUIP-Score, short for \textbf{Qu}oted \textbf{I}nformation \textbf{P}recision}
\newcommand{\overlapshort}{QUIP} 

\newcommand{\promptingstrat}{\emph{according-to}}
\newcommand{\dataportrait}{\textsc{Data Portrait}}

\definecolor{lightgreen}{RGB}{223,255,219}
\definecolor{lightred}{RGB}{255,219,219}
\definecolor{blueish}{RGB}{31, 78, 192}
\definecolor{orangeish}{RGB}{240, 147, 41}
\definecolor{mypink1}{rgb}{0.858, 0.188, 0.478}
\definecolor{mypink2}{RGB}{219, 48, 122}
\definecolor{mypink3}{cmyk}{0, 0.7808, 0.4429, 0.1412}
\definecolor{mygray}{RGB}{220, 220, 220}
\definecolor{darkbluee}{RGB}{0,17, 113}
\definecolor{purpleNew}{RGB}{151, 45, 204}
\definecolor{purplebg}{RGB}{229, 199, 244}
\definecolor{violet}{rgb}{0.70,0.05,0.65}

\usepackage{amsmath}
\usepackage[bb=dsserif]{mathalpha}
\def\mystrut(#1,#2){\vrule height #1pt depth #2pt width 0pt}

\makeatletter
\def\blfootnote{\xdef\@thefnmark{}\@footnotetext}
\makeatother

\newcommand\extrafootertext[1]{%
    \bgroup
    \renewcommand\thefootnote{\fnsymbol{footnote}}%
    \renewcommand\thempfootnote{\fnsymbol{mpfootnote}}%
    \footnotetext[0]{#1}%
    \egroup
}

\newcommand{\redtext}[1]{\colorbox{lightred}{\mystrut(.5, .5) #1}}
\newcommand{\greentext}[1]{\colorbox{lightgreen}{\mystrut(.5, .5) #1}}
\newcommand{\greytext}[1]{\colorbox{mygray}{\mystrut(.5, .5) #1}}

\newcommand{\myparagraphnoskip}[1]{\vspace{1mm}\noindent\textbf{#1}}

\title{\emph{``According to \ldots''}: Prompting Language Models \\  Improves Quoting from Pre-Training Data}

\author{Orion Weller$^*$, Marc Marone$^*$, Nathaniel Weir, \\
\textbf{Dawn Lawrie, Daniel Khashabi, Benjamin Van Durme} \\ 
Johns Hopkins University \\
\normalsize{\texttt{\{oweller2,mmarone1\}@jhu.edu}}
}

\begin{document}
\maketitle
\begin{abstract}
Large Language Models (LLMs) may hallucinate and generate fake information, despite pre-training on factual data. Inspired by the journalistic device of \emph{``according to sources''}, we propose \promptingstrat{} prompting: directing LLMs to ground responses against previously observed text.  
To quantify this grounding, we propose a novel evaluation metric (\overlap{}) that measures the extent to which model-produced answers are directly found in underlying text corpora. 
We illustrate with experiments on three corpora (Wikipedia, PubMed, and the U.S. legal tax code) that these prompts improve grounding under our metrics, with the additional benefit of often improving end-task performance.
Furthermore, prompts that ask the model to decrease grounding (or to ground to other corpora) indeed {decrease} \overlap{}, indicating the ability of LLMs to increase or decrease grounded generations on request.\footnote{We publicly release all code at \url{https://github.com/orionw/according-to}}
\blfootnote{* Authors contributed equally}
\end{abstract}

\section{Introduction}

As the deployment of Large Language Models (LLMs) in real-world applications continues to grow, their tendency to generate false content~\cite{ji-etal-2022-survey} poses significant risks to downstream users. Recent work has attempted to address this issue by augmenting them with retrieval~\cite{shuster2021retrieval,sun2022contrastive,borgeaud2022improving}; however, these models still struggle with hallucination problems in practice~\cite{liu2023evaluating}.

\begin{figure}[t]
    \centering
    \includegraphics[width=0.95\columnwidth,trim=0.6cm 0.2cm 0.4cm 0cm]{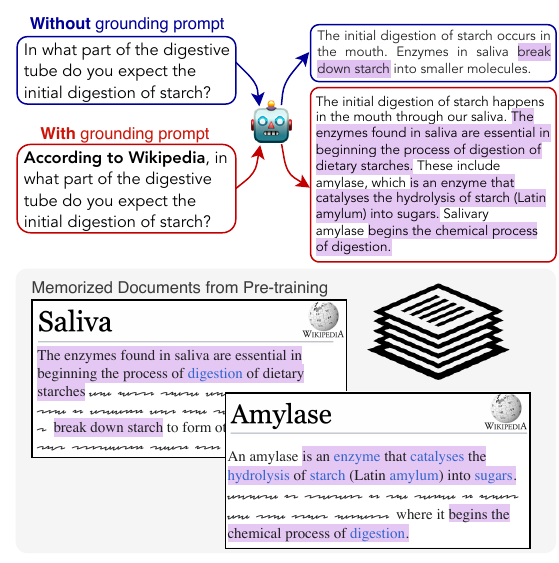}
    \caption{
   Prompting LLMs to respond with quotes directly from  pre-training data (shown in \colorbox{purplebg}{{purple}}).
   Prompting increases the proportion of quoted information.
    }
    \label{fig:teaser}
\end{figure}

This work explores the intriguing possibility of \emph{steering LLMs by prompting} them to quote 
more of the curated sources of information they have memorized during pre-training, thereby reducing their tendency to generate false information. 
As illustrated in \autoref{fig:teaser}, we explore whether adding phrases such as ``According to Wikipedia'' can guide LLMs to quote from Wikipedia, which is presumably observed in the pre-training corpus.
We find empirical evidence that this is attainable using current LLMs (both open and closed source).

Our study is inspired by two recent research areas. First, larger LLMs can be more effectively guided using natural language prompts~\cite{ouyang2022training,wan2023poisoning,ganguli2023capacity}. Second, as LLMs grow in size, their ability to remember facts and statements from pre-training improves~\cite{kandpal2022large,tirumala2022memorization,carlini2022quantifying,Carlini2020ExtractingTD}.
Thus, we seek to steer LLMs to use their memorization for a positive purpose: producing more grounded outputs.

A key step in this study is quickly determining whether generated outputs overlap significantly with pre-training data; i.e., efficiently performing membership testing via a \dataportrait~\cite{marone2023dataportraits}.
We design a new metric called \overlaplong{}, which builds on \dataportrait{}s and takes advantage of its speed and efficiency. \overlap{} then calculates n-gram overlap, quantifying how much of a passage is formed of spans that are exactly contained in the corpus.

To illustrate \emph{according-to} prompting, 
we perform experiments
based on the task of open-domain question answering (ODQA),  for which provenance-grounded answers are of particular importance.
We collect human-authored prompts designed to steer generations toward information grounded in our target corpora (Wikipedia, PubMed, and the U.S. legal tax code). We observe that across all human-authored prompts, we can increase the amount of overlap with the chosen corpora by 5-105\% while maintaining or even improving the downstream performance.
We show results across numerous datasets and models, including both open- and closed-sourced LLMs.

Interestingly, we also observe the opposite phenomenon -- it is possible to \textit{discourage} LLMs from grounding via prompts that either discourage grounding or encourage grounding to other corpora.
For example, we find this can decrease overlap with Wikipedia while lowering performance on downstream tasks that rely on Wikipedia content. 

We conduct scaling experiments on different model sizes, which indicate that as size increases, so does the effectiveness of our proposed approach.
This suggests that hallucinations may diminish with further scaling of instruction following LLMs.

In summary, we present \promptingstrat{} prompting, a simple and effective approach to improving an LLMs' ability to generate more factual information. Additionally, we introduce \overlap{}, an efficient metric for measuring groundedness of LLM generations against their pre-training corpus. We experiment with various prompting strategies across models, datasets, and scaling trends, and we find that \promptingstrat{} methods consistently improve groundedness under our introduced metric.

\section{Related Work}

\paragraph{Memorization in LLMs.}
Large language models have been observed to memorize their training data \cite[][among others]{Carlini2020ExtractingTD,chang2023speak}.
This is problematic when web-scraped training data contains sensitive personal data or low-quality information sources \cite{dodge2021documenting, luccioni2021s}.
However, it can be beneficial for models to memorize content from \emph{carefully curated and trusted corpora}, where careful de-duplication~\citep{lee2022deduplicating} and curation strategies~\citep{feng-etal-2022-automatic} can improve language model quality \citep{pile}.
Work on analyzing memorization has proposed measuring n-gram overlap against the first page of Google Search results as a proxy for memorization, using exact matches~\cite{Carlini2020ExtractingTD} and BLEU~\cite{levy-etal-2021-investigating}.

We measure quoting (and thus, memorization in closed-book generation settings) building off of \citet{marone2023dataportraits} who propose using membership testing tools that they label \dataportrait{}s.
As one implementation, they use a Bloom Filter \citep{bloom1970spacetime} for storing n-grams. We use this method for checking membership in a corpus as it allows us to build a fast, lightweight, and scalable metric for measuring quotation against large amounts of data (see Section~\ref{sec:quip} for details).

\paragraph{Hallucination and grounding.}
\label{sec:hallucinations}
Numerous studies~\cite{de-cao-etal-2021-editing, li-etal-2022-pre,weller2023generative} have demonstrated that LLMs struggle with both hallucination and factuality, leading to frequent inaccuracies and outright falsehoods. Previous research has attempted to alleviate this problem in various ways, including retrieving grounded documents before generation~\cite{sun2022contrastive,borgeaud2022improving,mallen2022trustlmretrieval,weller2022defending}, applying new decoding approaches~\cite{he2022rethinking}, post hoc tuning of LLMs~\cite{menick2022teaching,lee2022factuality}, and analyzing the model's output training data~\cite{han2022orca,park2023trak}.
Crucially, these works have a common thread: showing that grounding LLM generations results in fewer hallucinations \cite{lazaridou2022internet,Andriopoulos2023AugmentingLW}.
Our work focuses on a subset of grounding, \emph{quoting}, and is driven by the simple premise that \emph{anything quoted is grounded and not hallucinated.}
Our work therefore builds off the established research and is complementary to it, as we investigate a novel yet straightforward approach to steer LLMs towards more factual responses.

\paragraph{Attribution.}
A related line of work is attribution of generated text to their  sources~\cite{rashkin2021measuring,bohnet2022attributed}. 
Our work is related to this literature in that, our approach allows provable attribution to macro-level sources of information, such as Wikipedia or medical articles. 
However, we do not focus on offering any fine-grained 
attribution to the originating source documents. 
Given these distinctions our focus here is different  from --and complementary to-- the attribution literature.

\paragraph{LLM Steerability via prompting.} The larger LMs become, the easier they are to steer with natural language prompts \cite{kandpal2022large,carlini2022quantifying,mishra2022reframing,srivastava2023beyond}. Several works~\cite{mishra2022cross,chung2022scaling,wang2022benchmarking,wan2023poisoning} have shown that larger instruction-tuned models are more easily steered than smaller and non-instruction-tuned models. This is desirable in our setting, as we seek to use these capabilities of LLMs for a novel application of steerability: \emph{quoting more from a given corpus.}

\paragraph{Improving LLMs through prompting.}
Much recent work has focused on improving LLM performance on various benchmarks by improving the prompt given to the model. A sub-genre of these works includes those that ask the model to produce text before generating the answer, such as Chain-of-Thought \cite{wei2022chain} or Recitation-based Generation \cite{sun2022recitation}. We differ from these works by generating the answer first, then the explanation, indicating that our performance gains are not due to the same phenomena. Furthermore, our paper's focus is on improving LLM's ability to quote, rather than improving end-task performance.

\section{Methodology}
\paragraph{Defining Grounding}
There are many definitions of \textit{grounding} in the community \cite{bohnet2022attributed, mallen2022trustlmretrieval}. While acknowledging the broad scope of the term, we adopt a narrow definition: we call generated text \textit{grounded} with respect to a corpus if it is an exact quotation from the corpus. This is more stringent than some definitions because it does not count semantic grounding, e.g. when lexical forms do not match; however, quotation is one form of grounding that is intuitive and simple to measure.\footnote{We leave it to future work to expand our metric to the semantic grounding case, as semantic grounding (e.g. finding paraphrases) while matching the generations over an entire corpus is non-trivial; using retrieval systems biases the model towards lexical match (even for dense retrieval, c.f. \citet{macavaney2022abnirml}) and existing work in attribution/grounding does not scale to allow grounding to numerous (2+) passages.} Hence, we use  \textit{quoting} and \textit{grounded} interchangeably. 

\subsection{\overlap{}: Measuring Grounding to Pre-Training Data}
\label{sec:quip}
In order to understand grounding and quoting from model pre-training data, we need a metric to measure quoting. 
An intuitive approach is to use an n-gram measure, which can compare n-grams found in an LLM's generation to those in a corpus.
Such a quotation metric must be efficient to scale to large reference corpora. 

\paragraph{Problems with existing N-gram metrics} Existing n-gram metrics like BLEU or ROUGE store counts of n-grams from the references.
However, storing counts requires the use of data structures like a conventional hashtable, which is computationally difficult for a large corpus like Wikipedia. 
We estimate naively scaling \texttt{sacrebleu}~\cite{post-2018-call} to use Wikipedia as a reference would consume $\sim$1.5 TB of RAM (\autoref{app:bleu}).

\paragraph{\overlap{}} To enable efficient measurement of quoting from pre-training data, we start with a Bloom filter-based \dataportrait~\cite{marone2023dataportraits}, which allows for both faster and more memory efficient boolean membership queries than allowed by methods that use a hashtable to store counts. 
The Bloom filter approach enables one-time indexing of a large corpus with constant time lookups.

We define our new metric, \overlap{}, as the character n-gram precision of overlap between generated output and the pre-training corpus.\footnote{QUIP scores are not comparable across datasets, as they are specific to a given corpus. This is acceptable for our experiments that compare generations against one corpus.} More formally, for generation $Y$ and text corpus $C$:
\vspace{-0.2em}
\[
\text{\overlapshort{}}(Y; C) = \dfrac{\sum_{\substack{\text{gram}_n \in Y}} \mathbb{1}_{C}(\text{gram}_n)}{|\text{gram}_n \in Y|}, 
\vspace{-0.3em}
      \] 
where $\mathbb{1}(.)$ is an indicator function implemented with the \dataportrait{}: 1 if $\text{gram}_n \in C$ else 0.
Thus, a score of 0.5 would indicate that 50\% of the generated text $n$-grams are found in the pre-training corpus.
We macro-average this quantity over a set of generations to obtain a single performance number for a given test dataset.

\paragraph{\overlap{} Implementation} We build the \dataportrait{} on the version of Wikipedia included in the Pile,\footnote{ \href{https://www.tensorflow.org/datasets/catalog/wikipedia\#wikipedia20200301en}{wikipedia/20200301.en}} as it allows for us to exactly test the pre-training data included in many models like GPT-J%
\footnote{
Note, for several models evaluated here (e.g. OpenAI models) the exact Wikipedia version trained on is unknown.
} (See \S\ref{sec:other_corpora} for experiments applying \overlap{} to other corpora).
We use character-based n-grams as opposed to token-based, as different models have different tokenization schemes. Furthermore, character-based n-gram metrics have widespread usage in machine translation with metrics like chrF/chrF++ \cite{popovic2015chrf,popovic2017chrfpp}. We chose 25 character grams for the sketch\footnote{Not having multiple n-gram sizes like BLEU typically does allows us to significantly reduce memory consumption and had similar results to averaging across sizes.} (approximately 5 words) as we found it empirically gave meaningful results (neither too small nor too large an n-gram). 
Note that because the \dataportrait{} checks for exact matches it is sensitive to orthographic variation (e.g. case, whitespace), We view \overlap{} as a lower bound on actual quoting performance. 

\subsection{Validity of \overlap{}}
As \overlap{} is an n-gram metric, it inherits many of the same qualities of established metrics like BLEU and ROUGE. Further, many previous works have established the connection between higher amounts of grounding and fewer hallucinations (\S\ref{sec:hallucinations}). Building upon these previous studies, we establish that \overlap{} (1) accurately measures quoting like other n-gram metrics and (2) 
is correlated with fewer hallucinations.

We first conduct a straightforward experiment: what is the \overlap{} when measuring entirely quoted documents (e.g. exact Wikipedia pages) vs documents that are not necessarily quotes (e.g. from the Pile)? We randomly sample 1000 documents from each.  We find that the average \overlap{} for Wikipedia documents is 99.9\%\footnote{\overlap{} is 99.9 due to a single very short sampled document, where length < n-gram size} with a standard deviation of 0.1\% while on the Pile it is 17.0\% $\pm$ 0.8\%. Thus we can see that \overlap{} correctly measures full quotations and that random text has approximately 17\% 
\overlap{}.

\begin{table}[t]
    \small
    \centering
    \begin{tabular}{c|rrrr|r}
    \toprule
        \overlap{} & None & Some & Major. & All & Halluc. \\ 
        \midrule
        0.0 -- 0.25 & 12\% & 76\% & 12\% & 0\% & 20\% \\ 
        0.25 -- 0.5 & 0\% & 16\% & 84\% & 0\% & 22\% \\ 
        0.5 -- 0.75 & 0\% & 0\% & 80\% & 20\% & 12\% \\ 
        0.75 -- 1.0 & 0\% & 0\% & 48\% & 52\% & 6\% \\ 
        \bottomrule
    \end{tabular}
    \caption{Random sampled generations from NQ, binned by \overlap{}. \textbf{As \overlap{} increases, quoting increases and hallucinations decrease.} \textit{Major.} stands for Majority, while \textit{Halluc.} stands for Hallucination \%.}
    \label{tab:quip_bins}
\end{table}

Next, we consider partial, contextual quotations as found in LLM generations from NQ. We bin generations by \overlap{} ranges, sampling 50 from each bin. We then conduct two manual analyses: (1) how much of the generations are a quotation (none, some, majority, or all/nearly all) and (2) whether the generation is a hallucination (using gold provenances and answers, plus Google Search when unsure). 
\autoref{tab:quip_bins} shows that {as \overlap{} increases, the amount of quotations increases and the amount of hallucinations decreases.}

We do not expect these results to be surprising, as they have been demonstrated by a large amount of literature on n-gram metrics \cite{belz2006comparing,reiter2009investigation,popovic2015chrf}, and by the grounding and hallucination literature \cite{lazaridou2022internet,borgeaud2022improving,Andriopoulos2023AugmentingLW}. However, this analysis empirically demonstrates that using quoting for grounding and \overlap{} as the n-gram metric retains these desired properties.

\section{Grounding via \promptingstrat{} Prompting}
The previous results show 1) that we can efficiently measure quotation rate and 2) that more quotations correlate with fewer hallucinations.
Next, we seek to improve knowledge grounding by causing LLMs to quote directly from trusted resources seen during training.\footnote{Since we want to know what the LLM recalls on its own, we specifically do not use any retrieval models.}
We hope to access \emph{helpful} memorized content: strings copied from high-quality or trusted documents. 
We induce this behavior by taking a normal task prompt (e.g. an ODQA question) and appending an instructional phrase that encourages grounding such as ``Respond by using information from Wikipedia in your response".\footnote{We tried appending, prepending, and their combinations in early experiments and found that appending the grounding/anti-grounding prompts performed the best.}
We call this strategy \promptingstrat{} prompting.
Our experiments measure the change in \overlap{} of generations from a \promptingstrat{} prompt vs one without the extra instruction (i.e. a null prompt).

To verify that prompts can both increase and decrease grounding,
we also include prompts that are anti-grounding (e.g. ``Respond by using information from [another source] in your response" or ``Respond without using any information from Wikipedia.'')
This allows us to test the hypothesis that models can ground (or not ground) to a given corpus when asked because of the semantic meaning of the prompt, rather than the length of the prompt.
As prompting is notoriously brittle (e.g. changing the phrasing can affect the results) we provide a number of grounding and anti-grounding prompts to test whether these prompts provide consistent gains or are merely prompting artifacts (see \autoref{tab:chatgpt} for the list of prompts used).

{ %
\setlength\tabcolsep{5 pt} %

\newcommand{\chg}[1]{{\tiny(#1)}}
\newcommand{\up}{\cellcolor{lightgreen}}  
\newcommand{\dn}{\cellcolor{lightred}}  
\newcommand{\same}{\cellcolor{mygray}}  

\begin{table*}[]
\small	
\centering
\resizebox{1.0\textwidth}{!}{
\begin{tabular}{cp{9cm}|cccccccc}
\toprule
\multicolumn{2}{l}{\textbf{Prompt}}  & \multicolumn{2}{c}{\textbf{TQA}} & \multicolumn{2}{c}{\textbf{NQ}} & \multicolumn{2}{c}{\textbf{Hotpot}} & \multicolumn{2}{c}{\textbf{ELI5}} \\ 
\multicolumn{2}{l}{(appended after the question)} & \overlapshort & EM & \overlapshort & EM & \overlapshort & F1 & \overlapshort & R-L \\ 
\midrule
\multicolumn{2}{l}{$\emptyset$ (no additional prompt)} & 31.6 & 77.8 & 32.8 & 32.9 & 28.3 & 35.7 & 24.1 & 22.7 \\ 
\midrule
 \parbox[t]{2mm}{\multirow{8}{*}{\rotatebox[origin=c]{90}{
 \parbox[c]{3cm}{\centering \scriptsize grounding prompts}}}}
& \scriptsize "Based on evidence from Wikipedia:" & \dn 31.1 & \dn 77.3 & \same 32.8 & \up 34.0 & \dn 28.1 & \up 35.9 & \up 26.3 & \dn 22.3 \\ 
&  \scriptsize "As an expert editor for Wikipedia, I am confident in the following answer." & \up 31.7 & \dn 73.2 & \up 33.0 & \dn 30.2 & \up 28.7 & \dn 35.3 & \up 25.5 & \same 22.7 \\ 
&  \scriptsize "I found some results for that on Wikipedia. Here's a direct quote:" & \up 31.7 & \dn 70.1 & \up 33.8 & \dn 27.6 & \dn 28.1 & \dn 33.1 & \up 27.2 & \dn 21.0 \\ 
&  \scriptsize "Reference Wikipedia when answering the following question." & \up 32.8 & \dn 75.9 & \up 34.6 & \up 34.4 & \up 28.9 & \up 35.9 & \up 25.7 & \dn 22.0 \\ 
& \scriptsize "Answer according to Wikipedia." & \up 33.6 & \up 78.8 & \up 34.3 & \up 34.8 & \up 29.2 & \up 36.6 & \up 26.5 & \dn 21.7 \\ 
&  \scriptsize "Go to https://www.wikipedia.org and find direct quotes to answer the question.  Response: "" & \up 34.5 & \dn 72.7 & \up 32.9 & \dn 31.7 & \up 30.4 & \dn 35.5 & \up 25.8 & \dn 20.4 \\ 
& \scriptsize "Respond by using information from Wikipedia in your response." & \up  34.9 & \dn 76.3 & \up 35.3 & \same 32.9 & \up 29.9 & \up 36.1 & \up 26.3 & \dn 21.9 \\ 
&  \scriptsize "Respond to this question using only information that can be attributed to Wikipedia."& \up 35.7 & \dn 76.6 & \up 37.0 & \up 33.9 & \up 30.4 & \up 36.2 & \up 28.0 & \dn 21.5 \\ 
\midrule
\parbox[t]{3mm}{\multirow{3}{*}{\rotatebox[origin=c]{90}{
\parbox[c]{1cm}{\centering \scriptsize anti-grounding}}}} & \scriptsize "Respond by using information from Reddit in your response." & \dn 26.1 & \dn 75.8 & \dn 26.5 & \dn 31.6 & \dn 22.4 & \dn 35.0 & \dn 21.9 & \dn 22.2 \\ 
& \scriptsize "Respond by using information from Github in your response." & \dn 26.7 & \dn 74.3 & \dn 28.2 & \dn 32.4 & \dn 23.2 & \dn 33.7 & \up 24.3 & \dn 22.0 \\ 
& \scriptsize "Respond without using any information from Wikipedia in your response." & \dn 30.4 & \dn 76.9 & \dn 32.0 & \dn 32.0 & \dn 26.8 & \dn 32.9 & \up 24.7 & \dn 22.1 \\ 
\midrule
 \multicolumn{2}{l}{Zero-Shot No-Retrieval SOTA} & - & \textit{68.2} & - & \textit{24.9} & - & \textit{44.6} & - & \textit{22.7} \\
\multicolumn{2}{l}{Retreival-Augmented SOTA} & - & \textit{89.4} & - & \textit{60.4} & - & \textit{51.4} & - & \textit{26.5} \\
\bottomrule
\end{tabular}
}
\caption{
Impact of various prompts on the grounding (\overlap{}) and performance scores,  using ChatGPT (\S\ref{sec:main:result}). The top row is the null prompt (no additional prompt other than the question), the middle section includes grounding prompts, and the last section includes anti-grounding prompts.  We find that \textbf{grounding prompts generally improve the \overlap{} while anti-grounding prompts generally reduce \overlap{}}.
Colored cells indicate changes (\greentext{gains}, \redtext{losses}, or the \greytext{same}) relative to the null row. 
ELI5 ROUGE-L (R-L)
is based on similarity to Reddit rather than Wikipedia.
See \S\ref{app:sota} for sources of SOTA results.
}
\label{tab:chatgpt}
\end{table*}
}
\begin{table*}[t!]
\centering
\small
\begin{tabular}{l|rrrrrrrr}
\toprule
& \multicolumn{2}{c}{\textbf{TQA}} & \multicolumn{2}{c}{\textbf{NQ}} & \multicolumn{2}{c}{\textbf{Hotpot}} & \multicolumn{2}{c}{\textbf{ELI5}} \\ 
\multicolumn{1}{l}{\textbf{Model}}  & \overlapshort{} & EM & \overlapshort{} & EM & \overlapshort{} & F1 & \overlapshort{} & R-L \\ 
\midrule
Text-Davinci-003 & +14.7\% & +5.3\% & +14.7\% & +20.6\% & +14.4\% & +7.2\% & +16.5\% & -3.8\% \\ 
GPT-4 & - & - & - & - & - & -  & +17.6\% & -2.3\% \\ 
GPT-J Instruct & +12.1\% & - & +15.2\% & - & +13.9\% & - & +18.1\% & -2.5\% \\
Koala 7B & +5.1\% & - & +6.3\% & - & +5.0\% & - & +35.5\% & +14.6\% \\ 
FLAN-T5 XXL & +43.3\% & - & +41.5\% & - & +20.7\% & - & +105.2\% & +48.4\% \\ 
\bottomrule
\end{tabular}
\caption{Percent improvement of \textit{according-to} over null prompt. \textbf{The \textit{according-to} prompt improves performance in nearly every dataset and metric by 5-15\%}. 
We omit EM/F1 scores of smaller models for which 
our
prompting methods yield the same answer for grounding and null  (\S\ref{sec:models}). 
Due to cost, we only evaluate GPT-4 on ELI5.
}
\label{tab:all}
\end{table*}

\subsection{Datasets}
We use a variety of datasets to test if LLMs are consistent and to check whether grounding affects the end-task performance of a given dataset. 
To best measure the grounding of the output however, the model generations must be long enough to have many n-grams that can be measured.
Thus, we test on long-form question answering (QA), and for datasets that do not lend themselves well to long-form output (e.g. short-form QA) we ask the models to generate both the answer and a corresponding explanation whose n-grams can be measured.

Note that our purpose is not to improve state-of-the-art performance on these tasks, as our main research question is to analyze the grounding of model outputs. However, we note that \textit{according-to} prompting often achieves competitive or improved performance compared to other prompting baselines, as it naturally correlates with the ability to answer questions from the grounded material.

\noindent
We use the following datasets, each of which targets factual knowledge in Wikipedia:
\textbf{ELI5}~\cite{fan2019eli5} (the KILT~\citet{petroni-etal-2021-kilt} version), \textbf{Natural Questions}~\cite{kwiatkowski2019natural}, \textbf{TriviaQA (TQA)}~\cite{joshi2017triviaqa}, and \textbf{HotpotQA}~\cite{yang2018hotpotqa}. These datasets comprise a mixture of short- and long-form plus single- and multi-hop QA. \S\ref{app:datasets} provides further details.

\begin{figure*}[t]
    \centering
    \includegraphics[trim=10 10 0 0,clip,width=0.49\textwidth]{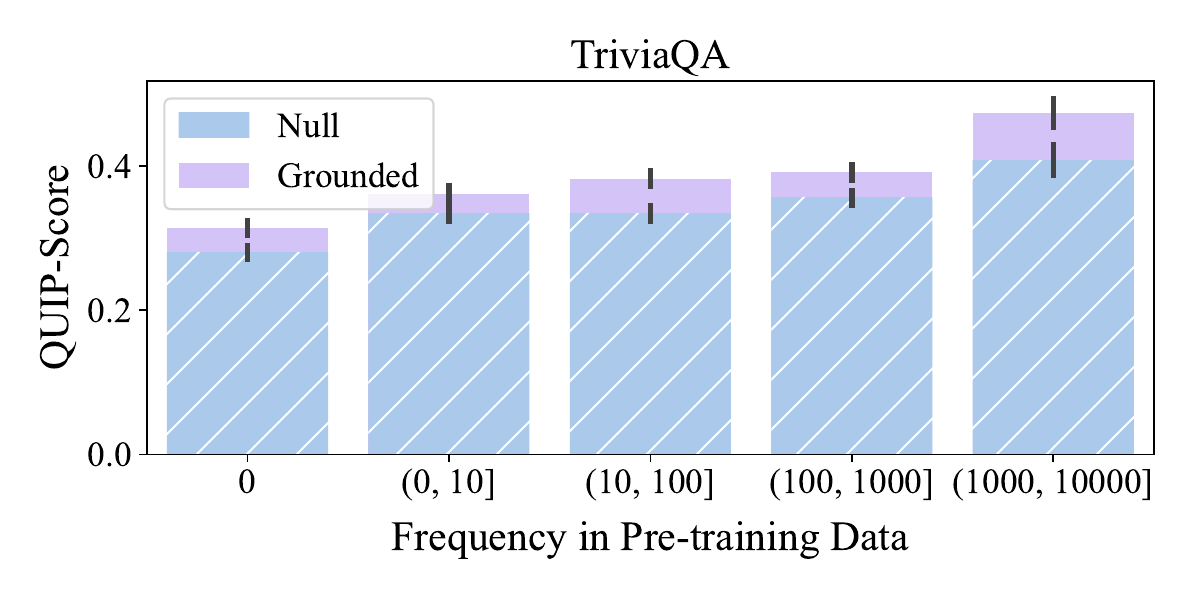}
    \includegraphics[trim=0 10 0 0,clip,width=0.49\textwidth]{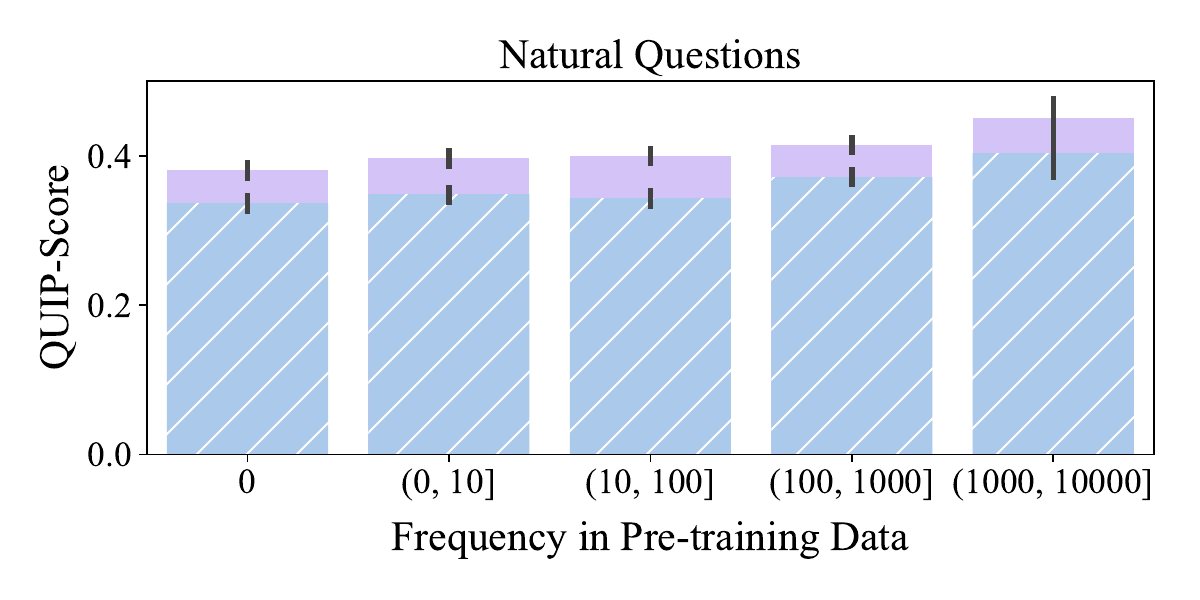}
    \caption{Impact of entity popularity on \overlap{}s, showing that \textbf{models are better able to quote pre-training text about popular entities}. The x-axis shows how many times the given entity relationship was found co-occurring in pre-training data. Bars indicate 1 standard error. We use the ranges following \cite{kandpal2022large}.}
    \label{fig:popularity}
\end{figure*}
\begin{figure}[htb!]
    \centering
    \includegraphics[trim=10 10 0 0,clip,width=0.47\textwidth]{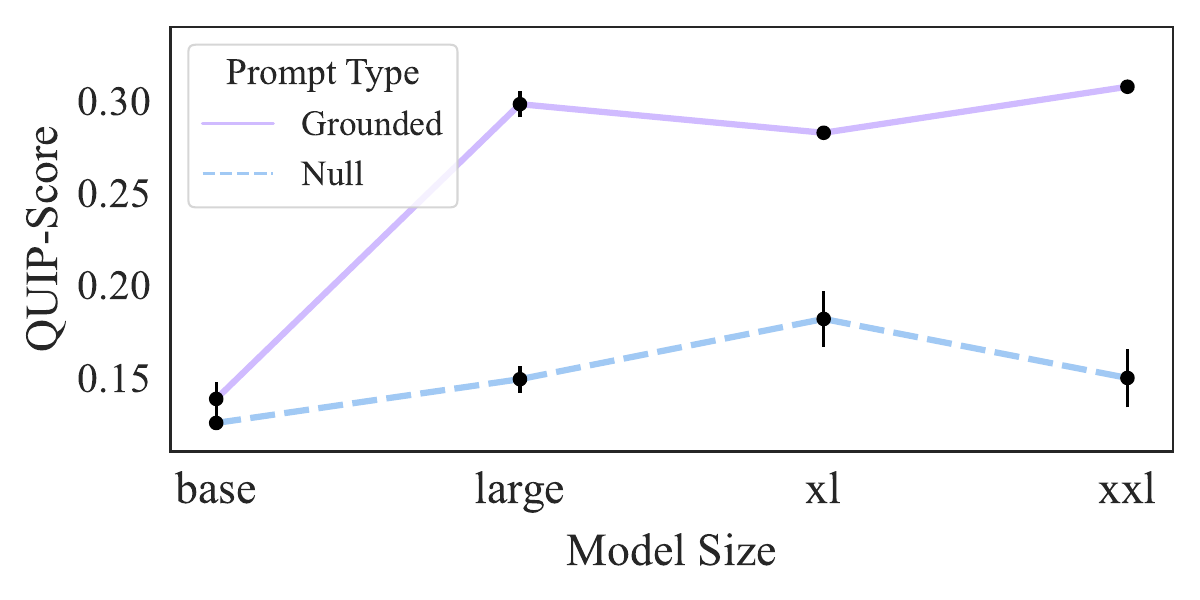} \\
    \includegraphics[trim=0 10 0 0,clip,width=0.48\textwidth]{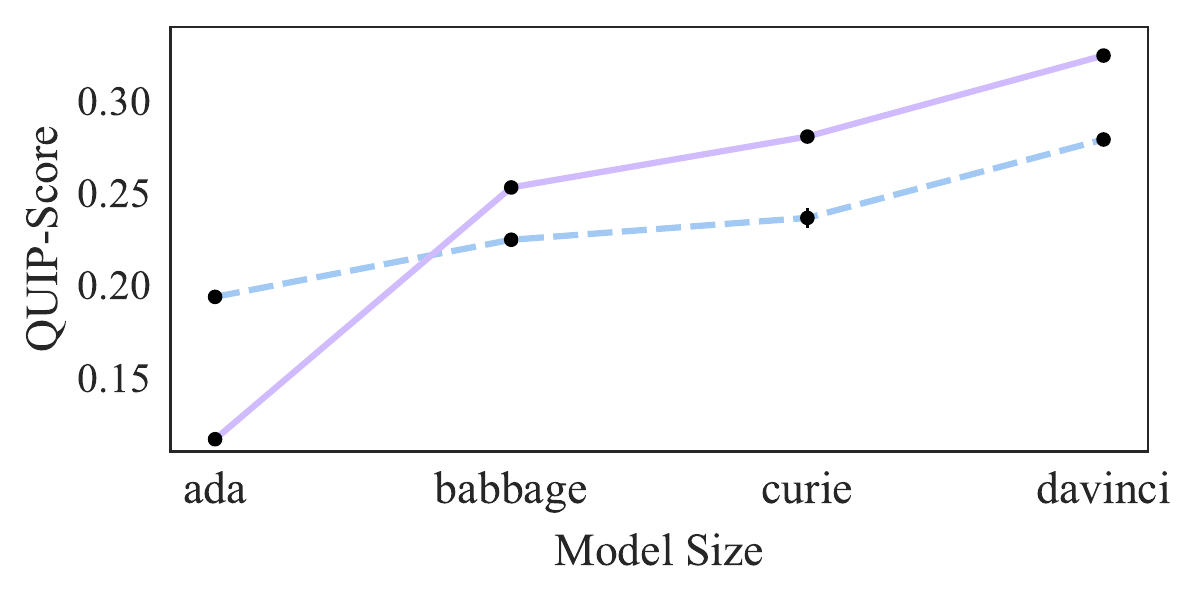}
    \caption{
    Model size vs \overlap{} performance using FLAN-T5 (top) and OpenAI (bottom) models. \textbf{As model scale increases, so does performance}. At smaller model sizes, the grounding prompt is not more effective than the null prompt, but gains efficacy with model size. Error bars indicate 1 standard error.\vspace{-0.28em}}
    \label{fig:scaling}
\end{figure}

\subsection{Models and Prompting}
\label{sec:models}
We test a wide array of models in our experiments including most OpenAI models~\cite{wang2023selfinstruct}, 
T5-based models (T5 adapted to language modeling,~\citealt{raffel2020exploring,lester2021power}
and FLAN-T5 \citealt{chung2022scaling}), GPT-J instruction tuned\footnote{\href{https://huggingface.co/nlpcloud/instruct-gpt-j-fp16}{https://huggingface.co/nlpcloud/instruct-gpt-j-fp16}} \cite{wang2021gpt}, and Koala \cite{koala_blogpost_2023} (a Llama variant, \citealt{touvron2023llama}). 
By doing so, we provide (1) results on both open and closed-source models, (2) results for models using many variations of instruction-tuning data, and (3) models ranging from 220 million parameters to 175B models. Note that our experiments consist solely of providing prompts to the models and do not include fine-tuning (as the goal is to see what these models can do zero-shot).

For short-form QA datasets, we prompt models to produce an answer plus an explanation, then measure \overlap{} of the latter. We found smaller models (e.g. < 15B parameters) were not able to follow instructions to provide both answer and explanation in a parseable format from just one prompt. Thus, we do two-step prompting with them, first for the answer, then for the explanation (and appending the grounding prompt, if used). 
\S\ref{app:full} provides prompting details and full text of the prompts used.

\section{Results}
\label{sec:main:result}
We first analyze a wide range of \promptingstrat{} prompts on ChatGPT.
We then test the null prompt and the best performing \promptingstrat{} prompt on a variety of other models for further analysis.
\autoref{tab:chatgpt} shows results for different prompts using ChatGPT.
There is a clear trend under which {all \promptingstrat{} prompts perform similarly or improve upon \overlap{} compared to the null.}
\overlap{}s for the anti-grounding prompts are the same or worse than the null prompt (i.e. no additional text) and significantly worse than the \promptingstrat{} prompts. 

Surprisingly, we find that \promptingstrat{} prompts also perform similarly, and sometimes even better than, the null prompt on end task performance (e.g. up to a 6\% improvement on NQ, 2.5\% on HotpotQA). This is not the case for ROUGE-L on ELI5, as that metric measures lexical similarity to Reddit, rather than similarity to Wikipedia.

We use these results on ChatGPT to inform our next experiments, using the null prompt and the best grounding prompt (``Respond to this question using only information that can be attributed to Wikipedia'') in our future experiments due to cost.

\subsection{Results from Other Models}
We show the relative difference of the grounding prompt over the null prompt for more models in \autoref{tab:all}, which further confirms our findings (for the absolute instead of relative numbers, see Appendix~ \ref{app:full}).
For example, using the grounding prompt with Text-Davinci-003 improves over the null prompt by around 15\% \overlap{} and 5-20\% for the specific task. For all models evaluated, the grounding prompt improves in both end-task performance and \overlap{} by 5-105\%.

Thus, our findings hold for a wide variety of models and model sizes -- even when prompts are not tuned for the specific model being prompted, indicating the generality of our approach.

\subsection{Impact of Model Size}
Does model size impact their ability to quote from their pre-training data? 
We answer this question using \overlap{} in \autoref{fig:scaling}, which shows that smaller models perform the same (for FLAN-T5 models) or worse (for OpenAI models) with a grounding prompt as opposed to the null prompt.
However, larger models perform significantly better with the grounding prompt as opposed to the null prompt, for both OpenAI models and FLAN-T5 models. We can conclude that a \textbf{model's ability to quote from its pre-training data improves with size}.

\begin{figure}
    \centering
    \includegraphics[trim=30 1cm 0 0,width=0.45\textwidth]{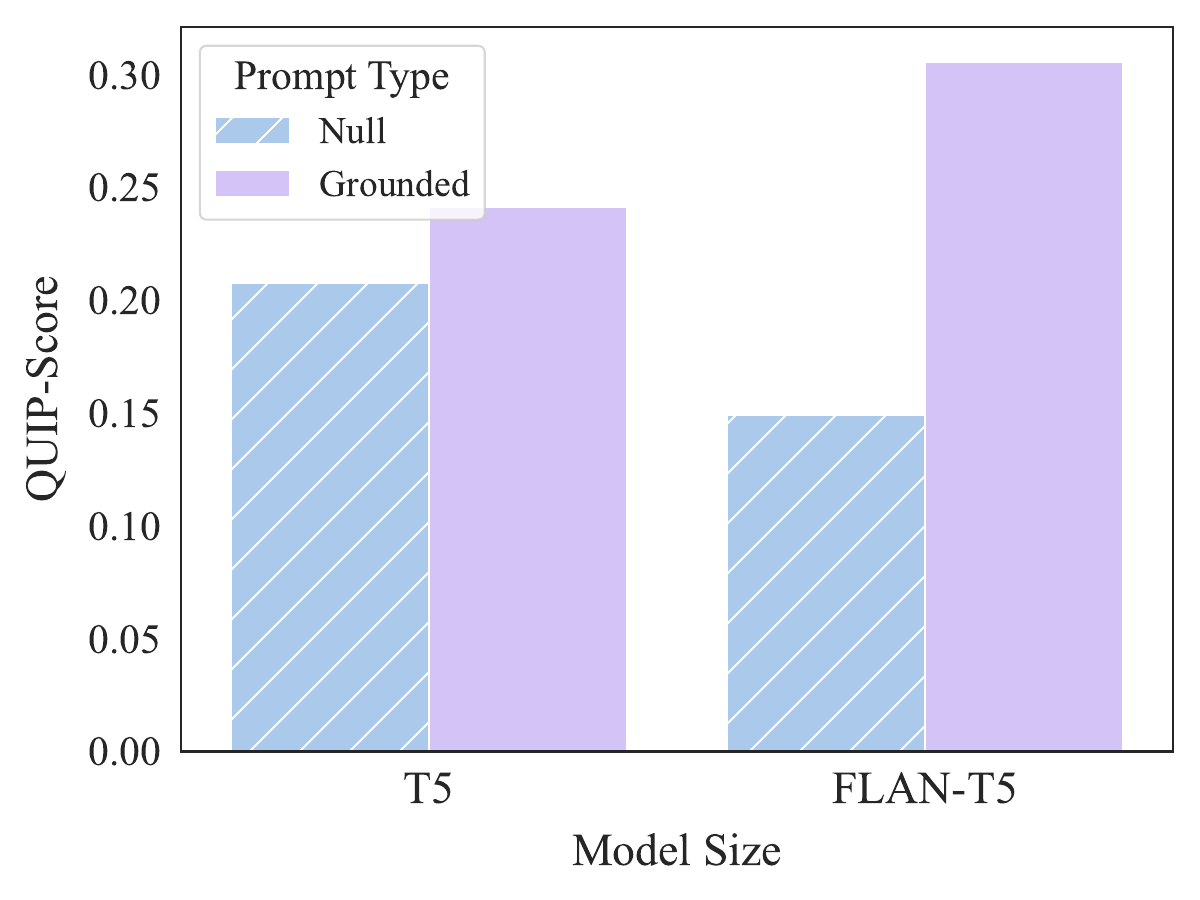}
    \caption{Comparing instructed-tuned FLAN-T5 XXL to non-instruction tuned T5-v1.1-Adapt XXL. Note that \textbf{grounding has a larger impact on instruction-tuned models as compared to non-instruction tuned}.}
    \label{fig:instruct}
\end{figure}

\subsection{Impact of Entity Popularity}
Another potential factor influencing generation of memorized content is the popularity of the entities mentioned in a question \cite{kandpal2022large,carlini2022quantifying}. 
Previous work has shown that entity co-occurrence (as measured by the number of times in the pre-training set that the entities in the question and in the answer co-occur in the same passage) is strongly correlated with task performance~\cite{kandpal2022large}.
We use their code and data (from the Pile) to explore whether \overlap{} correlates with co-occurrence frequency.

Due to the imbalance between co-occurrence counts, we sample 400 instances (or as many as available) from each dataset and co-occurrence frequency bin.\footnote{See \citet{kandpal2022large} for frequency bin design details.}
We measure the \overlap{} on these instances using the output generations from ChatGPT on both grounding and null prompts.

\autoref{fig:popularity} shows that QA entity popularity is positively correlated with \overlap{} for both grounding and null prompts, more so for grounding. We find that the model better recalls information from Wikipedia when QA entities frequently co-occur.

\subsection{Impact of Instruction Tuning}
One potential reason for why these models can recall their pre-training data on request is a better capability to instruction-follow. We test this hypothesis in Figure~\ref{fig:instruct} that compares T5-11B compared to FLAN-T5-11B. 

We find that instruction-tuning does help, as the \overlap{}s for T5-v1.1-Adapt are similar between grounding and null prompts, while the FLAN-T5 model has a large difference between the null and grounding prompt (roughly 2x better).

\subsection{Qualitative Examples}
\autoref{tab:examples} shows example outputs from a variety of models. 
When queried with \textit{according-to} prompts, the model generates much larger chunks of texts that occur in Wikipedia (shown in \textcolor{purpleNew}{{purple}}).

It is important to note that although the text may be \textit{grounded}, it does not mean the output generation is \textit{correct} with respect to the question. For example, the TriviaQA example shows that both models predicted the incorrect quote for Smokey the Bear, although the \promptingstrat{} explanation was better grounded in Wikipedia.

\section{Grounding to Other Corpora}
\label{sec:other_corpora}
In the previous experiments we've shown that LLMs can ground to Wikipedia upon request. However, are \promptingstrat{} prompts limited to the Wikipedia domain only? To answer this question we build two more \textsc{Data Portrait}s, one on PubMed articles and one with the U.S. legal tax code applying to tax year 2022. We then evaluate on datasets where having such background knowledge would be helpful and evaluate both the downstream task performance and \overlap{}, similar to the previous Wikipedia experiments.

{

\setlength{\tabcolsep}{4pt}

\begin{table}[t!]
\centering
\small
\vspace{0.5em}
\begin{tabular}{l|ll|ll}
\toprule
 & \multicolumn{2}{c|}{\textbf{Null}} & \multicolumn{2}{c}{\textbf{Grounded}} \\
 \multicolumn{1}{l|}{\textbf{Dataset}} & \overlapshort{} & EM &  \multicolumn{1}{c}{\overlapshort{}} & \multicolumn{1}{c}{EM} 
 \\ 
\midrule
\multicolumn{4}{l}{\textit{According to PubMed\dots}} \\ 
PubMedQA & 54.1 & 48.2 & 59.6$\uparrow$ (\textbf{+5.5}) & 49.6$\uparrow$ (\textbf{+1.4}) \\ 
MedQA & 45.1 & 53.3 & 45.9$\uparrow$ (\textbf{+0.8}) & 54.0$\uparrow$ (\textbf{+0.7}) \\
MedicationQA & 36.7 & N/A & 39.6$\uparrow$ (\textbf{+2.9}) & N/A \\ 
\midrule 
\multicolumn{4}{l}{\textit{According to the U.S. Tax Code\dots}} \\
SARA & 4.4 & 52.0 & 13.3$\uparrow$ (\textbf{+8.9}) & 55.0$\uparrow$ (\textbf{+3.0}) \\
\bottomrule
\end{tabular}
\caption{Results with ChatGPT using \textit{according-to} prompts for PubMed (top) and the U.S. legal tax code (bottom).
\textit{according-to} prompts consistently improve quoting on the non-Wikipedia domains while maintaining task performance. MedicationQA does not have an automated evaluation metric, so only QUIP is reported.
}
\label{tab:pubmed}
\end{table}

\begin{figure*}[!tb]
\scriptsize	
\centering
\renewcommand{\arraystretch}{1.2}
\setlength{\tabcolsep}{5pt}
\begin{tabular}{>{\raggedright}p{2.4cm}|p{12.9cm}} \toprule[1.5pt]
\textbf{Prompt} & \textbf{Generated Text} \\
\midrule
\textbf{1. Question (ELI5)} &  \textbf{Why is it that every time we put something down our throat toothbrush, finger etc. we start to gag but when we but food and water down our throat we don't gag at all?},
  \\
\midrule
Null Prompt \ \  (ChatGPT) & This is because our ga\textcolor{purpleNew}{{g reflex is triggered by}} objects that touch the back of our throat and activate \textcolor{purpleNew}{{sensory receptors called}} the pharyngeal receptors. When we eat or drink, the muscles in our throat automatically contract and move food and liquid down to the stomach, bypassing the pharyngeal receptors and avoiding the gag reflex. However, when we put non-food objects like a toothbrush or finger in our throat, they can \textcolor{purpleNew}{{activate these receptors and trigger the gag reflex.}} \\

Grounded Prompt (ChatGPT) & The gag reflex \textcolor{purpleNew}{{is a natural protective mechanism of the body that can be triggered by stimulation of the soft palate, the back of the tongue or the back of the throat. It is a reflex contraction of the muscles of the back of the throat}} that can cause vomiting or choking. The reflex can be more easily triggered by \textcolor{purpleNew}{{foreign objects, such as}} toothbrushes or fingers, \textcolor{purpleNew}{{that are not natural to the}} body, compared to food and water which are familiar substances that we regularly inge\textcolor{purpleNew}{{st without triggering the re}}flex. \textcolor{purpleNew}{{Additionally, the shape and texture of the foreign object may also contribute to triggering the}} gag reflex. \\

\midrule

\textbf{2. Question (NQ)} & \textbf{In what part of the digestive tube do you expect the initial digestion of starch?} \\
\midrule
Null Prompt \ \ (ChatGPT) & The initial digestion of starch occurs in the mouth; enzymes in saliva \textcolor{purpleNew}{{break down starch}} into smaller molecules.

 \\

Grounded Prompt (ChatGPT) & The initial digestion of starch happens in the mouth; S\textcolor{purpleNew}{{alivary amylase, an enzyme found in the mouth, is responsible for breaking down starch into simpler sugars}}.
 \\

\midrule

\textbf{3. Question (ELI5)} & \textbf{Why do towels that feel so rough on the skin dry you so much better, but towels that feel soft don't dry well at all?}\\

 \midrule
 
Null Prompt \ \ \ \\  (Davinci) & Towels that feel rough on the skin \textcolor{purpleNew}{{are usually made of a co}}arser fabric, such as terry cloth, which is made of loops of cotton yarn. This fabric is more absorbent than \textcolor{purpleNew}{{softer fabrics, such as}} velour or microfiber, which are made of finer fibers. The loops of terry cloth fabric \textcolor{purpleNew}{{create more surface area for the t}}owel \textcolor{purpleNew}{{to absorb water, which is why it is more effective at dr}}ying than softer towels.
 \\
Grounded Prompt (Davinci) & Towels that feel rough on the skin \textcolor{purpleNew}{{are usually made of a material with a higher absorbency}}, such as cotton or linen. T\textcolor{purpleNew}{{hese materials are able to absorb more water than softer materials, such as m}}icrofiber or polyester. The absorbenc\textcolor{purpleNew}{{y of the material allows it to draw moisture away from the skin more effectively, resulting in better d}}rying. Soft t\textcolor{purpleNew}{{owels, on the other hand, are usually made of materials with lower absorbency, which means they are not able to absorb as much water and therefore do not}} dry as well. \\

\midrule
 
\textbf{4. Question (TriviaQA)} &  \textbf{According to Smokey The Bear, Remember... only you can what?}\\
 
\midrule
 
Null Prompt (GPT-J-Instruct) & ``Only you can see me"; Because fire requires a fuel source to burn, and smoke is the byproduct of a fire's bu\textcolor{purpleNew}{{rning process. Therefore, s}}ince smoke is created whenever a fire burns, \textcolor{purpleNew}{{the presence of smoke in}}dicates that a fire is burning somewhere.
\\

Grounded Prompt (GPT-J-Instruct) & ``Only you can remember"; According to Smokey The Bear, memory \textcolor{purpleNew}{{is an internal mental process that is unique to each person. It is the ability to remember past experiences, along with the ability to recognize and interpret the information that is stored in our memories. This explains why, despite the fact that mem}}ory \textcolor{purpleNew}{{techniques vary between individuals, m}}emory is \textcolor{purpleNew}{{a universal quality that}} all humans share. \textcolor{purpleNew}{{Therefore, the answer to the question is}} ``yes".\\

\midrule

\textbf{5. Question (HotpotQA)} & \textbf{Which magazine is based out of more countries, Adventist World or Baseball Digest?}\\

\midrule
Null Prompt \ \ (FLAN-T5 XXL) & adventist world; Adventist World is a magazine that is based out of many countries.
 \\ 
Grounded Prompt (FLAN-T5 XXL) & adventist world; \textcolor{purpleNew}{{Adventist World is a monthly magazine published by the Seventh-day Adventist Church. It is based in the United States, Canada, and many other countries.}}
 \\
 
\bottomrule[1.5pt]
\end{tabular}
\caption{
Example generations from various considered models.
\textcolor{purpleNew}{Purple text was found in Wikipedia.}
Note that for non-ELI5 datasets, models were prompted to generate the answer, a semicolon, and then the explanation (see \S\ref{sec:models}). Note that better grounding to Wikipedia does not always imply correct answers (see Question 4).
}
\label{tab:examples}
\end{figure*}

\paragraph{Datasets.} We use datasets in the MultiMedQA benchmark suite \cite{singhal2022large} to test grounding to PubMed: PubMedQA \cite{jin2019pubmedqa} a reading comprehension task over PubMed abstracts, MedQA \cite{jin2020disease} consisting of multiple-choice questions from the US Medical Licensing Exam, and MedicationQA \cite{abacha2019bridging} which asks open-domain questions about patient medications. Although these last two are not directly sourced from PubMed, they contain information that is likely to be found in it. Note that we do not give the model the abstract as typically done in PubMedQA, but instead evaluate closed-book in order to measure quotes from model parameters.

In the legal domain, we use the SARA dataset \cite{holzenberger2020dataset} consisting of tax cases to be evaluated using natural language inference.\footnote{As these datasets have different formats, e.g. NLI and multiple choice, we change the prompt slightly to accommodate them (Appendix~\ref{app:additional}). We use the test set for all datasets.}

\paragraph{Results.} The results in ~\autoref{tab:pubmed} with ChatGPT show that \promptingstrat{} prompts improve end-task performance and \overlap{}s. On SARA, \overlap{}s more than triple, while also minorly increasing performance. In the medical domain, grounding to PubMed improves performance slightly as well, and improves QUIP scores on all datasets.

\section{Discussion and Future Implications}
Our results strongly suggest that LLMs can be steered via prompting to increase the amount by which they quote human-authored sources in their training data. This finding has strong implications not just for our considered tasks, but also for a wide array of other task spaces in which provenance grounding is important.

We note that our \promptingstrat{} prompting strategy is orthogonal to other directions in LLM grounding, including using retrieval augmentation, and as \promptingstrat{} prompting is simple and generally increases both grounding and task performance we would encourage future research to try our approach in tandem.

\section{Conclusion}
Large language models struggle with hallucination, or generating incorrect information, despite the large amount of factual pre-training data they were trained on. 
To help alleviate this problem, we proposed \textit{according-to} prompts, asking language models to ground their output to their pre-training corpus.
To quantify the extent to which models achieve this goal, we introduced a new metric, \overlap{}, that efficiently and quickly measures the percent of the model's generation that exists as exact quotes in the pre-training corpus.
We showed that prompting models with \promptingstrat{} prompts greatly improves the \overlap{} while anti-grounding prompts reduces the \overlap{}, across a variety of domains and corpora.
Our analysis also shows that \overlap{} increases with the popularity of the entity in the question and the model size.
We hope that this work brings more attention to the positive aspects of LLM memorization and encourages more work into understanding LLM grounding to their pre-training data.

\section{Limitations}
Our proposed metric only accounts for exact lexical match and will miss other types of grounded statements - thus we view \overlap{} as a lower bound on grounding where grounding is defined only by quoting from source material. \overlap{} is also \dataportrait{} specific, as the amount of n-grams in the portrait affect the scores. We leave it to future work to generalize this metric, as our work focuses on using it to compare two prompts with the same Portrait.

We also recognize the possibility of a discrepancy between the pre-training data of private models like ChatGPT and the Wikipedia version we use for analysis, due to limited information on their pre-training. However, this might not be a significant concern, as although Wikipedia is not completely static, a substantial part of the information in this knowledge source remains consistent over a short span of years. Furthermore, our results with ChatGPT are similar compared with models for which we do have the exact pre-training data (like GPT-J).

\section*{Acknowledgements}
This work has been supported in part by the U.S. National Science Foundation under grant No. 2204926. OW and NW are also supported by the National Science Foundation Graduate Research Fellowship Program.

\bibliography{anthology,custom}

\appendix

\section{Dataset Details}
\label{app:datasets}
\myparagraphnoskip{ELI5}, or ``Explain Like I'm 5'' \cite{fan2019eli5} is a long-form QA dataset composed of user questions and answers from the subreddit r/ELI5. 
We use the KILT version \cite{petroni2021kilt} dev set of ELI5 since it is a ``grounded'' subset of the original (with the non-grounded questions filtered out), allowing a more suitable evaluation of our research question.

\myparagraphnoskip{Natural Questions (NQ)}~\cite{kwiatkowski2019natural} is a short-form (< 5 word answer) QA dataset gathered from real-world Google searches. To compare with previous work in prompting on NQ, we evaluate on the full development set.

\myparagraphnoskip{TriviaQA (TQA)}~\cite{joshi2017triviaqa} was collected by scraping question and answer pairs from trivia websites, and then matching the answers (short-form) to Wikipedia passages. Following previous work, we use the filtered dev set (7k instances).

\myparagraphnoskip{HotpotQA}~\cite{yang2018hotpotqa} is a multi-step short-form question-answering dataset that requires two-step reasoning to come to the correct answer. It was gathered from crowdsourcing questions and answers from Amazon Mechanical Turk using two-hop links on Wikipedia. We use the full dev set.

\section{All Results}
\subsection{Sources for SOTA Performance in \autoref{tab:all}}
\label{app:sota}
SOTA zero-shot results are from LLaMA 33B and 65B \cite{touvron2023llama}, PaLM 540B \cite{wang2022rationale}, and BART \cite{Su2022ReadBG} respectively. For retrieval-augmented SOTA, we show \citet{izacard2022atlas} for NQ, TriviaQA and HotpotQA, and \citet{Su2022ReadBG} for ELI5.\vspace{1em}

\subsection{Additional Models for \textit{According-To} vs Null}
We show all results for the models in Table~\ref{tab:all} that did not fit due to space. 

\paragraph{Prompts for Short-Form QA.} For short-form QA datasets, to help the models generate both the answer and the explanation in a parseable format, we append the following prompt before the question:

\begin{quote}
    You are a highly intelligent \& complex question-answer generative model. You take a question as an input and answer it by imitating the way a human gives short answers with a corresponding explanation. You answer should be short - only a few words.\textbackslash{n}\textbackslash{n}Your output format should be the answer, then a semicolon, then the explanation.\textbackslash{n}
\end{quote}

For models that don't respond well to the above prompt (or similar prompts aimed at generating both answer and explanation from one generation), we use the following prompts in a two step manner:

\begin{quote}
    Output the answer only. \{Insert Question\}\textbackslash{n}Answer string only:
\end{quote}

\begin{quote}
        Question: \{Insert Question\}\textbackslash{n}Answer: \{Insert Previous Output\}\textbackslash{n}\textbackslash{n}\textbackslash{n}Give a detailed explanation for why this is true. \{Insert Grounding Prompt Here\} \textbackslash{n}Explanation:
\end{quote}

\paragraph{Prompt for ELI5 for smaller models.}
For T5-v1.1-Adapt and GPT-J-Instruct evaluated on ELI5, we append ``Answer:'' following the end of both the normal null and grounding prompts because otherwise the model outputs very short (< 5 words) or nonsensical responses. With this addition the model produces normal fluent text.

\label{app:full}
\begin{table*}[]
\small	
\centering
\begin{tabular}{ll|cccccccc}
\toprule
\multicolumn{1}{l}{\textbf{Model}} & \multicolumn{1}{l}{\textbf{Prompt}} & \multicolumn{2}{c}{\textbf{TQA}} & \multicolumn{2}{c}{\textbf{NQ}} & \multicolumn{2}{c}{\textbf{Hotpot}} & \multicolumn{2}{c}{\textbf{ELI5}} \\ 
     & & \overlapshort & EM & \overlapshort & EM & \overlapshort & F1 & \overlapshort & R-L \\ 
\midrule
Text-Davinci-003 &  Null & 35.9 & 68.2 & 38.7 & 24.3 & 34.6 & 29.2 & 27.7 & 23.7 \\ 
Text-Davinci-003 & Grounded & 41.2 & 71.8 & 44.4 & 29.3 & 39.6 & 31.3 & 32.2 & 22.8 \\ 
\midrule
GPT-4 & Null & - & - & - & - & - & - & 21.0 & 21.5 \\
GPT-4 & Grounded & - & - & - & - & - & - & 24.7 & 21.0 \\

\midrule
GPT-J-Instruct & Null & 28.1 & 2.2 & 28.2 & 0.9 & 29.2 & 7.0 & 22.8 & 19.9 \\ 
GPT-J-Instruct & Grounded & 31.5 & 2.1 & 32.5 & 1.0 & 33.2 & 7.0 & 27.0 & 19.4 \\  
\midrule
Koala & Null & 34.0 & 17.2 & 36.1 & 6.3 & 33.9 & 13.2 & 24.1 & 19.9 \\ 
Koala & Grounded & 35.8 & 17.2 & 38.4 & 6.3 & 35.6 & 13.2 & 32.6 & 22.8 \\ 
\midrule
FLAN-T5 XXL & Null & 18.6 & 31.5 & 23.5 & 13.3 & 25.8 & 23.6 & 14.9 & 12.4 \\ 
FLAN-T5 XXL & Grounded & 26.6 & 31.5 & 33.2 & 13.3 & 31.1 & 23.6 & 30.6 & 18.7 \\
\bottomrule
\end{tabular}
\caption{Full results for other models. Note that the low EM scores for GPT-J and Koala are due to the model failing to output short answers zero shot (e.g. ``the answer is...'' instead of outputting the answer. Both used the same instruction tuning dataset.). GPT-4 was only run on ELI5 due to cost. \label{tab:davinci}}
\end{table*}

\section{Conventional N-Gram Metrics}
\label{app:bleu}
Toolkits such as \texttt{sacrebleu} implement multiple n-gram metrics~\cite{post-2018-call}. However, these tend to use conventional data structures such as python sets and dictionaries.
These are not suitable for measuring n-gram metrics against very large references (i.e. the entirety of Wikipedia).
In \autoref{tab:set-sizes} we compare the sizes of several datastructures on a sample of $\sim100$M n-grams (approximately 0.07\% of the 25 char-grams in Wikipedia). 
The typical CPython set or dictionary implementation uses a hashtable of pointers to data elements (i.e. character n-grams or strings).
It requires substantial memory to store both the hashtable backing array and the string data elements (11,107 MiB). 
This could be optimized by storing only the table and not the data elements, introducing false positives for hash collisions (note that this is similar to a Bloom filter with $k=1$ hash functions).
One could also store only pointers (references) into the original text rather than copies of the string. 
These options are still larger than an optimal Bloom filter which uses around 14 bits per element for our chosen parameters.
On the sampled data, this consumes only 163 MiB of memory. 
Extrapolating these storage costs indicates that using a naive, un-optimized python set or dictionary would consume around 1.5TB of memory to store all n-grams. 

Note that these measurements are only for a single n-gram width.
If comparing \overlap{} to a metric that stores n-grams of multiple widths, this could further increase memory usage. 

\begin{table}[h!]
\centering
\begin{tabular}{@{}lr@{}}
\toprule
\textbf{Structure}        & \textbf{Size (MiB)} \\ \midrule
set              & 11107      \\
set (no elements) & 4096       \\
Bloom filter D.P & 163        \\ \bottomrule
\end{tabular}
\caption{Sizes of structures holding 100M n-grams.}
\label{tab:set-sizes}
\end{table}

\section{Prompts for Non-Wikipedia Datasets}
\label{app:additional}
We use the same style of prompt as in the Wikipedia sections, but modify them to adapt to the format of the tasks. For example, on the SARA dataset ChatGPT would predict entailment for every question unless additional wording was given to be more balanced in its prediction.

\paragraph{SARA.}
 \begin{quote}
     You are a highly intelligent \& complex legal statutory entailment system that checks whether a particular judgement holds for a particular case in the U.S. legal tax code. You take a the ruling of a legal situation and respond with a reply of contradiction or entailment, along with a corresponding two paragraph explanation. You answer should be short - only contradiction or entailment. Be sure to verify that the entailment is 100\% correct, otherwise choose contradiction.\textbackslash{n}\textbackslash{n}Your output format should be the answer, then a semicolon, then the verbose explanation.\textbackslash{n}\textbackslash{n}Premise:\{Insert Text Background\}\textbackslash{n}Hypothesis\{Insert Question\}\textbackslash{n}\textbackslash{n}Fill in the following:\textbackslash{n}ANSWER HERE; EXPLANATION HERE
 \end{quote}

\paragraph{PubMed.}
We use the same prompt as usually specified for Short Form QA above, except using 
{\tt ``\textbackslash{n}\textbackslash{n}According to PubMed,'' }
as the grounding prompt.

\paragraph{MedQA.}
We use the  same Short Form answer beginning prompt and the following changes to the grounding prompt, 
{ \tt ``\textbackslash{n}\textbackslash{n}According to PubMed the multiple choice answer is:\textbackslash{n}''} as without the multiple choice specifier it would fail to correctly predict the multiple choice answer.

\paragraph{MedicationQA.} We only append a grounding prompt and no prompt before the question (as it is not short form). We use the same grounding prompt as in PubMedQA.

\section{Computational Resources}
We use model APIs for experiemnts with OpenAI and use 1 A100 GPU for experiments with local models. Each experiment took less than an hour for each dataset approximately.

\end{document}